\documentclass[10pt,twocolumn,letterpaper]{article}
\usepackage{cvpr}
\usepackage{times}
\usepackage{epsfig}
\usepackage{graphicx}
\usepackage{amsmath}
\usepackage{amssymb}

\usepackage{multirow}
\usepackage{diagbox}
\usepackage[pagebackref=true,breaklinks=true,letterpaper=true,colorlinks,bookmarks=false]{hyperref}

\newcommand{\tabincell}[2]{\begin{tabular}{@{}#1@{}}#2\end{tabular}}

\cvprfinalcopy 

\def\cvprPaperID{6} 

\ifcvprfinal\fi

\begin{document}

\title{Deeply Exploit Depth Information for Object Detection}

\author{Saihui Hou, Zilei Wang, Feng Wu\\
	CAS Key Laboratory of Technology in Geo-spatial Information Processing and Application System, \\
	University of Science and Technology of China, Hefei, 230027, China\\
	{\tt\small  saihui@mail.ustc.edu.cn, \{zlwang, fengwu\}@ustc.edu.cn} \\
}

\maketitle
\thispagestyle{empty}

\begin{abstract}
This paper addresses the issue on how to more effectively coordinate the depth with RGB aiming at boosting the performance of RGB-D object detection. Particularly, we investigate two primary ideas under the CNN model: property derivation and property fusion. Firstly, we propose that the depth can be utilized not only as a type of extra information besides RGB but also to derive more visual properties for comprehensively describing the objects of interest. So a two-stage learning framework consisting of property derivation and fusion is constructed. Here the properties can be derived either from the provided color/depth or their pairs (\textit{e.g.} the geometry contour adopted in this paper). Secondly, we explore the fusion method of different properties in feature learning, which is boiled down to, under the CNN model, from which layer the properties should be fused together. The analysis shows that different semantic properties should be learned separately and combined before passing into the final classifier. Actually, such a detection way is in accordance with the mechanism of the primary neural cortex ($\text{V}_1$) in brain. We experimentally evaluate the proposed method on the challenging dataset, and have achieved state-of-the-art performance.
\end{abstract}

\section{Introduction}
\label{introduction}
The RGB-D images have been provided in the real-world visual analysis systems thanks to the wide availability of affordable RGB-D sensors, \textit{e.g.} the Microsoft Kinect. Compared with the primitive RGB, the RGB-D can bring remarkable performance improvement for various visual tasks due to the access to the depth information complementary to RGB~\cite{shotton2013real, fu2015object, kong2015bilinear}. Actually, the depth has some profitable attributes for visual analysis, \textit{e.g.} being invariant to lighting or color variations, and providing geometrical cues for image structures~\cite{socher2012convolutional}. For object detection, which is one of typical complex visual tasks, the acquisition of RGB-D images is applicable and beneficial. However, how to effectively utilize the provided depth information of RGB-D images is still an open question.

In recent years, Convolutional Neural Network(CNN) has achieved great success in computer vision and obtained the best performance in various visual tasks~\cite{szegedy2014going, hariharan2014simultaneous,long_shelhamer_fcn}. CNN is generally considered as an end-to-end feature extractor to automatically learn discriminative features from millions of input images~\cite{krizhevsky2012imagenet}. In this paper, we also adopt CNN to extract rich features from the RGB-D images, \textit{i.e.} we are under the CNN model to investigate the exploitation of the depth information.

\begin{figure}
\centering
\includegraphics[width=90mm]{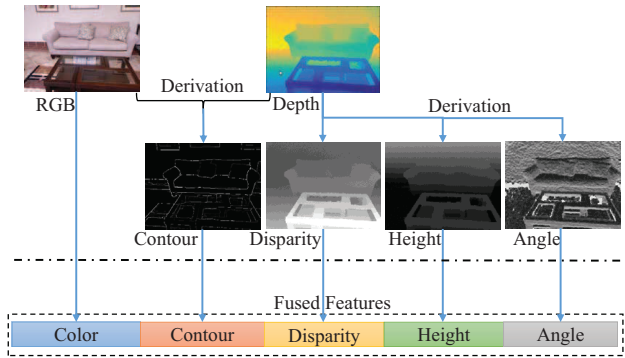}
\caption{
\textbf{Illustration of learning rich features for RGB-D object detection}.
Various property maps are derived to describe the object from different perspectives. The features for these maps are learned independently and then fused for the final classification. Specifically, the derived maps include geometry contour from the color/depth pairs, and horizontal disparity, height above ground, angle with gravity from the depth data. These maps, as well as the RGB image, are sent into different CNNs for feature learning. And the features are joint before being fed into the classifier.
}
\label{fig_concatenate_features}
\end{figure}

For the RGB-D object detection with CNN, the key is how to elegantly coordinate the RGB with depth information in feature learning. In the previous literatures, some intuitive methods have been proposed~\cite{couprie2013indoor,gupta2014learning}. Roughly, we can divide them into two broad categories according to the strategy the depth is treated. The first one is to straightforwardly add the depth map to CNN as the fourth channel along with the RGB~\cite{couprie2013indoor}. That is, the depth is processed in the same way as the RGB, and they are together convolved for granted. However, it makes no semantic sense to directly merge the depth and color maps, since they contain disparate information.
The second is to process the color and depth separately, and they are combined  before being fed into the final classifier, where the extracted features are joint. Specifically,  two independent CNN networks are learned: one for RGB and one for depth~\cite{gupta2014learning}. As for the depth network, the input can be the original depth data or encoded data from the depth, \textit{e.g.} height above ground, and angle with gravity~\cite{gupta2014learning}. It has been empirically shown that the second way usually outperforms the first one.
In this paper, we further investigate how to deeply exploit the depth information with aims of boosting the detection performance.

Before introducing our proposed method, we review the primary mechanism of human visual systems.
First, multiple visual properties are always used together to describe one object when people try to recognize it,
\textit{e.g.} geometry contour, color, and contrast~\cite{einhauser2003does}. And it is usually thought that exploiting more properties is much helpful. Second, the primary visual cortex ($\text{V}_1$),  which consists of six functionally distinct layers and is highly specialized in pattern recognition~\cite{gazzaniga2013cognitive}, abstracts different visual properties independently in the low layers and integrates in the relatively high layers.

Inspired by the working mechanism of $\text{V}_1$ area, we propose a novel method to deeply exploit the depth information for object detection. Figure.~\ref{fig_concatenate_features} illustrates the main ideas of our method. Firstly, various visual property maps are derived through analyzing the provided color and depth pairs. It is believed that more properties can contribute to the accurate description of the object and thus help boost the detection performance. Specifically, the derived properties include the contour, height, and angle maps\footnote{Indeed, other properties can be also adopted, which may be obtained by specific sensors or more advanced derivation methods. Considering the simplicity, only several directly computable properties are employed here.}. Secondly, we systematically investigate the method to fuse different visual properties under the CNN model, \textit{i.e.} how to represent a property, and from which layer the properties need to be fused together. The result of our analysis shows that the multiple properties should have complete and independent semantics in accordance with the human cognition, \textit{e.g.} RGB channels should be treated as a whole to represent the color property rather than separate them from each other, and it is better to fuse the different properties after they have been explicitly transformed into the high-level features.

We evaluate the proposed method on the challenging dataset \textit{NYUD2}, and the experimental results show that our method outperforms all the baselines and achieves state-of-the-art performance.


\section{Related Work}
\label{sec_related work}
Object detection~\cite{everingham2010pascal} is to mark out and label the bounding boxes around the objects in an image. Particularly, the adopted features are critical in determining the detection performance~\cite{sermanet2013pedestrian}. Traditional methods, including the MRF~\cite{krahenbuhl2012efficient} and DPM~\cite{felzenszwalb2010object}, are all based on the hand-crafted features such as SIFT~\cite{lowe2004distinctive} and HOG~\cite{dalal2005histograms}. However, these features are difficult to adapt to the specific characteristics in a given task. And more recent works~\cite{szegedy2013deep, erhan2014scalable, girshick2014rich} have turned to the Convolutional Neural Network (CNN), which can learn discriminative features automatically from millions of RGB images.
A typical CNN consists of a number of convolution and pooling layers optionally followed by the fully connected layers~\cite{krizhevsky2012imagenet}, and is able to learn multi-level features ranging from edges to entire object~\cite{zeiler2014visualizing}.

For object detection with CNN,  a proposed method is to build a sliding-window detector and then take the CNN for classification, which is usually applied on constrained object categories, \textit{e.g.} faces~\cite{rowley1998neural} and pedestrians~\cite{sermanet2013pedestrian}. And the object detection is formulated as a regression problem in~\cite{szegedy2013deep, erhan2014scalable} then the CNN is involved in predicting the localization and labels of the bounding boxes.
The most remarkable work lies in~\cite{girshick2014rich}. The system called R-CNN first generates around 2000 category-independent region proposals for an input image and then computes features of each region with the CNN. A category-specific SVM is appended to predict the label and score for each proposal.

When it comes to the RGB-D object detection, \cite{gupta2014learning} proves that CNN can also be trained to learn depth features from the depth map. Indeed, the extra depth exactly makes it easy to recognize human pose~\cite{shotton2013real}, align 3D models~\cite{gupta2015aligning}, and detect objects~\cite{socher2012convolutional, couprie2013indoor, gupta2014learning}.
Under the CNN model, two typical methods for RGB-D object detection have been proposed about how to utilize the depth information~\cite{couprie2013indoor, gupta2014learning}. One is to directly add the fourth channel for depth, and then equally convolve all channels in one network~\cite{couprie2013indoor}. The other is to separately process the depth and color (RGB) using two independent networks~\cite{gupta2014learning}. Obviously, these works mainly focus on the extraction of depth features, rather than considering thoroughly how to better coordinate the color and depth pairs for accurately describing the objects.

A more related work is the one by Gupta~\etal~\cite{gupta2014learning}, in which the depth data is encoded to three channels of horizontal disparity (D), height above ground (H), angle with gravity (A), and then form the DHA image into CNN to learn depth features, besides the RGB network. In our work, differently, the D, H, A are derived as new maps describing the object from different perspectives, and used to separately learn particular types of features encoding the multiple visual properties. More than that,  we propose to use the depth combining with the RGB to derive new maps to provide extra information, \textit{e.g.} the geometry contour. Furthermore, we systematically investigate the fusion way of different properties under the CNN model. We believe that our proposed detection framework and the investigation of feature fusion would inspire more advanced works to significantly improve the performance of object detection.

\begin{figure*}
\centering
\includegraphics[width=180mm]{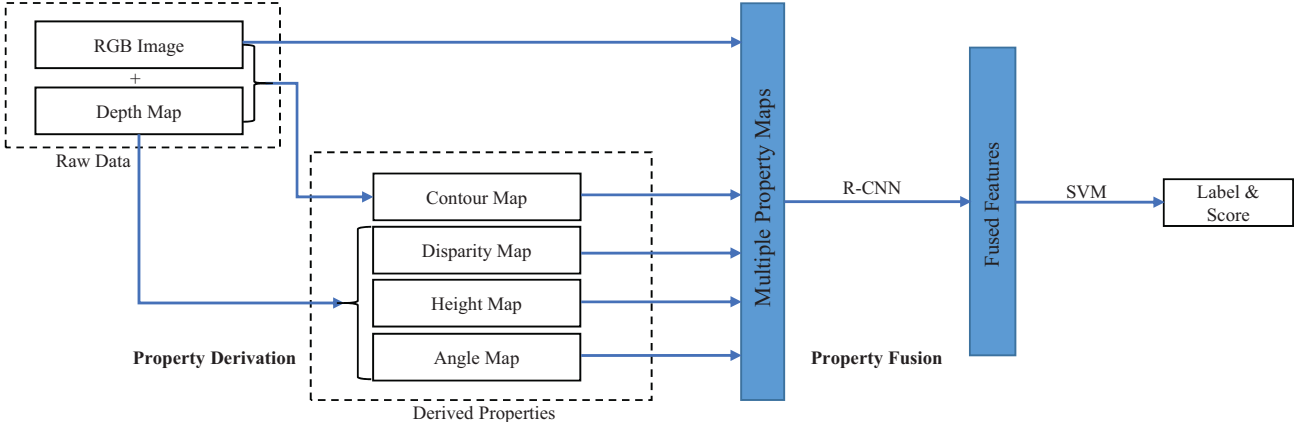}
\caption{
\textbf{Overview of our framework for RGB-D object detection}. We manage to learn multiple properties of the object and fuse them better for the detection. Various maps including geometry contour, horizontal disparity, height above ground and angle with gravity are first derived from the raw color and depth pairs. These maps, along with the RGB image, are sent into different CNNs to learn particular types of features. And the features are fused at highest level, \textit{i.e.} they are not joint until passing into the classifier.
The region proposals for R-CNN~\cite{girshick2014rich} are generated using MCG~\cite{arbelaez2014multiscale} with the depth information. A SVM is appended to predict the label and score for each proposal.
}
\label{fig_approach}
\end{figure*}

\section{Our Approach}
\label{sec_approach}

Intuitively, acquiring more information contents about the object can contribute to the more accurate recognition. Meanwhile, for human being, the visual cortex of the brain is exactly to abstract various types of visual information from the input scenes in the inception phase~\cite{gazzaniga2013cognitive}.
Inspired by such a principle, more informational sources are always desired in computer vision for high performance, \textit{e.g.} developing more powerful or precise sensors~\cite{shotton2013real}. In practical applications, however, the accessible sources are rather limited because of the constraints on deployment and cost. In this paper, we attempt to mine as much useful information as possible by analyzing thoroughly the available data, for the sake of boosting the performance of RGB-D object detection, \ie only the color image of single view and the corresponding depth map are originally provided.

To this end, we propose a novel two-stage feature learning framework for object detection on the RGB-D data, as shown in Figure~\ref{fig_approach}. Specifically, we first derive more property maps from the input color and depth pairs. This procedure functionally emulates the abstracting mechanism of primary visual cortex. These properties describe the object from multiple views, and combine with the raw data to form a relatively complete set of feature maps. Then we adopt the well-performed Convolutional Neural Network (CNN) model to generate image representations from these feature maps. Specially, the method to fuse different properties under the CNN model would be investigated systematically in this work.
Finally, we feed the joint representations into the SVM for classification. In the proposed framework, the property derivation and property fusion are especially important to determine the performance of object detection.

\subsection{Property derivation}
It is unlikely for any of existing methods, including the CNN, to learn the various properties of the objects accurately from the same input scene as the human brain does, especially when only limited data is available. So it is essential to derive more property maps complementary to the original color and depth. For human visual system, the geometric properties usually play an important role in recognizing the object, \eg shape and outline. Fortunately, the availability of depth information makes it possible to compute the geometric properties with greater accuracy,  compared to only using the plain color data. In this paper we mainly focus on the properties that can be derived from the raw color and depth data directly.

Specifically, the adopted properties include Ultrametric Contour Map (UCM), Horizontal Disparity (D), Height Above Ground (H), and Angle With Gravity (A). In particular, UCM representing the geometry contour is calculated using both the color and depth data, and the other three properties are only from the depth map. The horizontal disparity is an encoded version of the original depth map since it is more suitable for feature leaning with CNN~\cite{gupta2014learning}. Along with the color image, the final property maps comprise of RGB, UCM, D, H, and A. Now we elaborate on the computation of \emph{geometry contour}, \emph{height above ground} and \emph{angle with gravity}.

\subsubsection{Geometry contour}

In philosophy, \emph{geometry contour} tells directly the boundaries between the objects of interest and the background~\cite{dollar2013structured}. In object detection with CNN, however, the input data is only the images of rectangular area, which are produced by either sliding windows or region proposals~\cite{girshick2014rich}. That is, both the objects and context in bounding boxes are processed blindly. So explicitly providing the contour map would help CNN to mark out the objects more accurately. In this work, we particularly adopt the Ultrametric Contour Map (UCM) produced by the gPb-ucm algorithm~\cite{amfm_pami2011} combining the RGB and depth data.

\subsubsection{Height}

\emph{Height above ground} indicates the position of the objects \wrt different categories~\cite{gupta2013perceptual}, \eg a television is usually put on a table and a pillow is on the bed or sofa. The height map exactly represents such relations between the objects and thus is beneficial to distinguish the objects of interest. In this work, we produce the height map in an approximate way. Specifically, the 3D point cloud is first gained by processing the depth map. Then the height value of each pixel is roughly calculated by subtracting the lowest point (with the minimum height) within an image, which can be regarded as a surrogate for the supporting ground plane.

\subsubsection{Angle}

\emph{Angle with gravity} gives a lot of cues about the image structures and what the real world looks like, \eg the surface of a table or bed is usually horizontal while the wall or door is vertical. Such information provides important clues to the shape of the objects, and thus would be helpful for the object detection. In this work, we adopt the method in~\cite{gupta2013perceptual} to calculate the angle map.

Specifically, we first estimate the direction of gravity (DoG), denoted by $g$, using the depth data. In practice, $g$ is updated in an iterative manner and initialized with the Y-axis. For the current DoG ($g_{i-1}$), the aligned set $\mathcal{N}_{||}$ and orthogonal set $\mathcal{N}_{\perp}$ are produced with a angle threshold $d$, \ie
\begin{eqnarray}
&&\mathcal{N}_{||} =\{ n:  \theta_{n, g_{i-1}} < d\ | \ \theta_{n, g_{i-1}} > 180^{\circ} - d \}, \\
&&\mathcal{N}_{\perp} = \{ n: 90^{\circ} - d < \theta_{n, g_{i-1}} < 90^{\circ} + d \},
\end{eqnarray}
where $n\in \mathcal{N}$ represents the local surface normal, and $\theta_{n, g_{i-1}}$ denotes the angle between $n$ and $g_{i-1}$.
Then a new $g_i$ is estimated by solving the following optimization problem,
\begin{eqnarray}
\min \limits_{g_i:||g_i||=1}  \sum_{n \in \mathcal{N}_{||} } \sin^2\theta_{n, g_i} + \sum_{n \in \mathcal{N}_{\perp} } \cos^2\theta_{n, g_i}.
\end{eqnarray}
It means the estimated gravity is expected to be aligned to the normals in $\mathcal{N}_{||}$ and orthogonal to the ones in $\mathcal{N}_{\perp}$.

Once the DoG is obtained, we assign each pixel with the value of angle made by its local surface normal with $g_i$. Consequently, the angle map is formed with the same size as the original image.

\subsection{Property fusion}
\label{sec_fuse_features}

So far we have obtained multiple visual property maps to represent the rich information of the object, which include the color (RGB), geometry contour (UCM), horizontal disparity (D), height above ground (H), and angle with gravity (A). For object detection, an unified image representation is always needed for the final classification that integrates the useful information from all properties. For each property represented by one map (except the color with three channels of RGB), the sophisticated learning model can be directly applied to produce the corresponding feature, \eg CNN. But how to coordinate multiple different properties in feature learning is not fully explored yet. For example, an intuitive method is to straightforwardly input all property maps together into CNN with multiple channels~\cite{couprie2013indoor}, but the resulting performance may be unsatisfying. Thus the key of generating the joint discriminative image representations is to determine an integration method, which can better fuse the multiple properties.

In this section, we attempt to deeply investigate the ways to fuse the different properties under the CNN model. Particularly, part of property maps obtained in the previous stage would be directly adopted when the numerical analysis is needed.

\subsubsection{Fusion analysis}
\label{sec_fusion_analysis}
CNN can naturally learn the hierarchical features by different layers~\cite{zeiler2014visualizing}, in which the input map is finally transformed into a feature vector through a series of operations represented by convolution and pooling~\cite{krizhevsky2012imagenet}. When multiple property maps of the same scene are fed into CNN, we can view these maps as the lowest level features. Then the property fusion is essentially to determine that, from which layer different property features should be arithmetically calculated together.

There are two extreme cases for property fusion. One is to directly increase the number of channels in the input layer of CNN and accept all property maps equivalently. The other is to separately learn different property features with independent networks and concatenate them before passing into the final classifier. In this work, both cases are denoted by \emph{input} and \emph{final} respectively for convenience. Of course, we can also fuse different properties from certain middle layer, which actually means that different property features are learned independently before that layer and then drawn from the synthesis of properties in the subsequent layers.

\begin{figure}
\centering
\includegraphics[width=90mm]{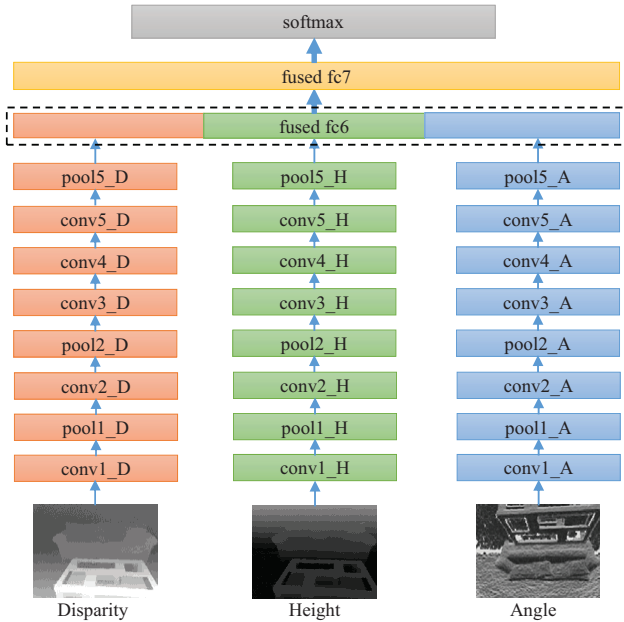}
\caption{
\textbf{The architecture of network when fusing from \emph{fc6}}. The D, H, A networks are independent before the first fused layer (\emph{fc6}) and initialized with the individual trained model. The \emph{fused fc6} and  \emph{fused fc7} are fully connected.
}
\label{fig_fused}
\end{figure}

We first analyze the underlying reasons for the difference of fusing the properties in multiple layers, and the classical AlexNet~\cite{krizhevsky2012imagenet} is adopted here. Figure~\ref{fig_fused} provides an exemplar of network architecture for fusion, where the properties are fused from the \emph{fc6} layer, and the three property maps of D, H and A are used without loss of generality.  For a specific fusion network, \eg fusing from \emph{fc6} here, the learned features in the previous layer are straightly stacked into multiple feature maps, and then these maps are processed without distinction in the subsequent layers. It can be observed that the inter-layer connections between different property features are added after the first fused layer (\emph{fc6}), which formally increases the network parameters. However, it is still not clear that how these connections impact on the final representations and also on the detection performance.

The typical CNN model has three major functional components, \ie \emph{convolution}, \emph{rectified linear units(ReLU)}, and \emph{max-pooling}. In particular, the \emph{convolution} belongs to the linear operators and thus theoretically has no great impact on fusing from different layers.
In contrast, both the \emph{ReLU} and \emph{max-pooling} are of high nonlinearity. So it is naturally supposed that different fusion schemes are through the \emph{ReLU} and \emph{max-pooling} to impact the learning results. We conduct an evaluation experiment in order to show the effect of three CNN components more intuitively. Here the dataset of \textit{NYUD2}~\cite{Silberman:ECCV12} is used, and the performance is evaluated on the \emph{val} set (See Section.~\ref{sec_experiments} for the details of dataset and model training). It is noticeable that we measure the impact of functional components by comparing the resulting performance of the two extreme fusion schemes (\ie \emph{input} and \emph{final}). In addition, the PCA is conducted on the joint image representations in \emph{final} scheme to keep the same dimension of classification features. The results are shown in the Table~\ref{tab_nyud2_fuse_analysis}.

\begin{table}[!tbp]
	\caption{\textbf{The performance on NYUD2 \emph{val} set to analysis the effect of \emph{ReLU} and \emph{max-pooling} on fusion}. For fusion schemes of \emph{input} and \emph{final}, the architecture of CNN for feature learning are respectively \emph{full}, \emph{conv+pool}, \emph{conv}, \emph{conv+relu}. The input are DHA image and D+H+A maps. The results are all mean $AP^b$ in percentage. See Section.~\ref{sec_fusion_analysis}.}
	\label{tab_nyud2_fuse_analysis}
	\begin{center}
		\begin{tabular}{|l|c|c|c|}
			\hline
			\diagbox{Arch}{Input} & \tabincell{c}{DHA\\ \emph{(input)}} & \tabincell{c}{D+H+A\\ \emph{(final)}}
			& \emph{gap}\\
			\hline
			\emph{full} & 24.75 & 28.85   &4.10    \\
			\hline
			\emph{conv+pool} & 12.42 & 14.52  & 2.10 \\
			\hline
			\emph{conv} & 1.58 & 1.95  &  0.37     \\
			\hline
			\emph{conv+relu} & 13.91 & 20.56 & 6.65  \\
			\hline
		\end{tabular}
	\end{center}
\end{table}
\begin{table}[!tbp]
	\caption{\textbf{The performance on NYUD2 \emph{val} set when fusing features from different layers.} We fuse the D, H, A properties with the schemes of \emph{input}, \emph{final}, \emph{fc7}, \emph{fc6}, \emph{fc5} in turn. And it can be seen that the \emph{final} scheme performs best. The results are all mean $AP^b$ in percentage. See Section.~\ref{sec_fusion_experiment}.}
	\label{tab_nyud2_fuse_layers}
	\begin{center}
	\begin{tabular}{|l|c|c|c|c|c|}
		\hline
		fused level & \emph{final} & \emph{fc7} & \emph{fc6} & \emph{pool5}	& \emph{input} \\
		\hline
		mean $AP^b$(\%) & 28.85 & 26.03 & 25.99 & 23.61  & 24.75\\
		\hline
	\end{tabular}
	\end{center}
\end{table}

According to Table~\ref{tab_nyud2_fuse_analysis}, for the \emph{full} network (AlexNet), the detection performance of fusing with \emph{final} scheme is obviously higher than that of fusing with \emph{input}. Then the \emph{ReLU} (also normalization and dropout) is removed (\emph{conv+pool}), the gap goes down but still exists. In the conditions that only the \emph{convolution} is adopted\footnote{The \emph{fully connected layer} can be seen \emph{convolution} layer in some way. And for networks without \emph{max-pooling} layers, including the \emph{conv} and \emph{conv+relu}, the crop size for input images is 71*71.} (\emph{conv}), the mean performance of the detection for both fusion schemes are approximately equal. But the performance gap is widened when the \emph{relu} is imposed (\emph{conv+relu}). In summary, it is the nonlinear functional components (\ie \emph{ReLU} and \emph{max-pooling} for CNN),
that make the fusion schemes achieve a wide range of detection accuracy. Thus it is necessary to systematically explore the property fusion schemes and choose the one that performs best.

\subsubsection{Fusion scheme}
\label{sec_fusion_experiment}

In this subsection, we explore the property fusion schemes under the CNN model, \ie from which layer the different property features need to be processed jointly. To address such an issue, we particularly adopt the  strategy of experimental evaluation here, due to the intractableness of theoretical analysis. Specifically, we conduct multiple experiments using different fusion schemes, each of which represents starting the fusion from certain layer. We use the same experimental settings as in the previous subsection.
The working mechanism of visual cortex in the brain inspires that the properties should be fused in the high level. So here we only investigate several relatively high layers, \ie fusing from \emph{fc7}, \emph{fc6} and \emph{pool5} are particularly chosen along with the \emph{input} and \emph{final} schemes. Table~\ref{tab_nyud2_fuse_layers} reports the detection performance of different fusion schemes.

From Table~\ref{tab_nyud2_fuse_layers}, it can be seen that the \emph{final} scheme results in the highest detection accuracy. Fusing from \emph{fc6} and \emph{fc7} reach better performance than \emph{input}. But when fusing from \emph{pool5} a little accuracy decrease occurs, which may be caused by introducing too much network parameters in the middle layers, while only limited number of training samples are provided.
For the current architecture of CNN, therefore, it is recommended to fuse different visual properties in the final step, \ie the learning models for encoding the property maps into the features should be trained separately.  In the following experiments, the fusion scheme of \emph{final} is directly adopted.

\section{Experimental Result}
\label{sec_experiment_result}

In this section, we evaluate the proposed method on the challenging dataset \textit{NYUD2}~\cite{Silberman:ECCV12}, which is widely used in the previous works on RGB-D object detection~\cite{couprie2013indoor, gupta2014learning, gupta2015aligning}. It consists of 1449 images of indoor scenes with vast variations of clutter and noise, and is split here following the standard way: 795 images for \emph{trainval} set and 654 images for \emph{test} set. The 795 images are further divided into 381 for \emph{train} set and 414 for \emph{val} set as in~\cite{gupta2014learning}. The furniture categories that appear frequently in actual life are considered, such as bed, chair, door, table, television and so on. And the box detection average precision (denoted by $AP^b$~\cite{hariharan2014simultaneous}) adopted in the PASCAL VOC Challenges is taken as the metric to measure the detection performance.

\subsection{Model details}
\label{sec_experiments}
In our experiments, we consider the typical CNN architecture described in~\cite{krizhevsky2012imagenet} and its Caffe implementation~\cite{jia2014caffe} for object detection, as in the previous works~\cite{girshick2014rich, hariharan2014simultaneous, gupta2014learning}. And the \emph{liblinear}~\cite{fan2008liblinear} for SVM is adopted.
Our model training mainly consists of two aspects: finetune the CNNs for property feature learning, and train the SVM for proposal classification.

\textbf{Finetune CNN:}
There are five CNNs to train all together in our final system. They are absolutely independent with each other and can be trained in parallel. The derived D, H, A and UCM maps, all have the same size with the raw RGB image. They are linearly scaled to 0-255 range and replicated three times to match the architecture of CNN.

For method evaluation, we first finetune the networks on the \emph{train} set and report performance on the \emph{val} set.
We start the finetuning with the Caffemodel pretrained on ILSVRC12 dataset. The learning rate is initialized to $10^{-3}$ and decreased by a factor of 10 every 20k iterations. And the finetuning lasts for 30k iterations. Region proposals that overlap with the ground truth by more than 50\% are taken as positives, and labelled with the maximal overlapping instance's class, while the rest proposals are all treated as background. When it comes to the \emph{test} set,
the CNNs are trained again on the \emph{trainval} set. The learning rate is also set to $10^{-3}$ at first and decreased by a factor of 10 every 30k iterations and there are 50k iterations in all.

Note that there is no data augmentation in our model. After the finetuning is done, we cache and concatenate the features from \emph{fully connected layer 6 (fc6)} of each network for SVM training.

\textbf{Train SVM:}
The training is started with the hyper-parameters $C = 0.001, B = 10, w_1 = 2.0$. The positive set of each category is fixed to the ground truth boxes for the target class in each image, and the negative set is the boxes which overlap less than 30\% with the ground truth instance from that class.
The SVM is trained on features from the \emph{train} set for method evaluation on the \emph{val} set and trained on features from the \emph{trainval} for the \emph{test} set performance. At test time, non-maximum suppression with the threshold $0.3$ is first carried out on the proposals for each image and then perform evaluation.

\subsection{Method evaluation}
\label{sec_result_nyud2_val}
In Table~\ref{tab_nyud2_val}, we report the performance in $AP^b$ on the \emph{val} set for method evaluation. It was implied in Section~\ref{sec_fuse_features} that learning the features of different properties separately can achieve better performance. Here we further validate that with the complete property set.
We start from the model in~\cite{gupta2014learning}. The depth map is encoded with the three channels of DHA images for depth feature learning, and then combine with the color features from RGB network to get the result mean $AP^b$ of 31.43\% in Column I. It's the best performance on the \emph{val} set as we know before this paper (a little lower than~\cite{gupta2014learning} because we work with no data augmentation).

The first experiment is to separate the D, H, A maps into three networks (called DHA Separation for short) to train and cache features respectively, and then combine with the color features to get the mean $AP^b$ of 35.48\% in Column II. This procedure gives us a 4.05\% improvement (31.43\% to 35.48\%, 12.89\% relative), which is much surprising. However, we notice that the dimension of joint features after the DHA Separation increase by three times that from the DHA network.
In Column III, we take dimension reduction using PCA on the concatenated features from D, H, and A networks, to keep the dimension the same, and then combine with the color features too. We get the mean $AP^b$ of 34.45\%, which is still obviously higher than that before the DHA separation (31.43\% in Column I). It proves again that learning different property features independently is indeed more powerful.


\begin{table}[!tbp]
\scriptsize
\caption{\textbf{Control experiments on NYUD2 val set for RGB-D object detection.}
The difference among these methods mainly lies the input maps. ``+'' between different maps means they are sent into different CNNs for feature learning.
See Section.~\ref{sec_result_nyud2_val}}
\label{tab_nyud2_val}
\begin{center}
\begin{tabular}{|r|c|c|c|c|c|c|c|}
\hline
                & I & II & III & IV & V            \\
\hline
\tabincell{r}{input\\for CNN}
& \tabincell{l}{RGB+\\DHA} & \tabincell{l}{RGB+\\D+H+A} & \tabincell{l}{RGB+\\D+H+A}
& \tabincell{l}{R+G+B\\+DHA} & \tabincell{l}{RGB+D+\\H+A+UCM}
\\
\hline
PCA             & no & no & yes  & no & no                   \\
\hline
bathtub         & 19.90	& \textbf{36.76}	& 36.70	& 22.03	& 33.93
                       \\
bed             & 64.67	& 66.31	& \textbf{66.84}	& 61.64	& 63.95
                        \\
bookshelf       & 13.40	& 12.07	& 10.46	& 12.43	& \textbf{15.02}
                        \\
box             & 2.14	& 2.35	& 2.81	& \textbf{3.75}	& 2.56
                        \\
chair           & 39.15	& \textbf{44.99}	& 43.71	& 36.98	& 44.83
                        \\
counter         & 34.32	& \textbf{40.89}	& 40.51	& 38.82	& 40.13
                       \\
desk            & 10.94	& \textbf{11.50}	& 10.37	& 6.31	& \textbf{11.50}
                       \\
door            & 19.77	& 20.87	& 19.78	& \textbf{22.71}	& 18.71	
                        \\
dresser         & 23.91	& 23.97	& 23.98	& 19.47	& \textbf{26.69}	
                        \\
garbage-bin     & 37.19	& 42.31	& 41.89	& 32.65	& \textbf{42.52}
                        \\
lamp            & 35.51	& \textbf{39.95}	& 39.58	& 32.05	& 39.42
                       \\
monitor         & 41.84	& 42.15	& 41.14	& 41.62	& \textbf{44.23}
                         \\
night-stand     & 33.69	& 38.58	& 38.64	& 38.26	& \textbf{43.56}
                        \\
pillow          & 32.18	& 35.93	& 33.72	& 32.89	& \textbf{38.18}
                         \\
sink            & 34.86	& 42.45	& 42.14	& 38.36	& \textbf{43.47}
                         \\
sofa            & 39.88	& 45.72	& 45.57	& 40.04	& \textbf{47.70}
                         \\
table           & 17.77	& \textbf{24.20}	& 21.15	& 17.94	& 23.21
                        \\
television      & 44.46	& \textbf{37.16}	& 35.58	& 41.26	& 34.75
                        \\
toilet          & 51.62	& 65.97	& 59.94	& 51.40	& \textbf{69.45}
                        \\
\hline
mean            & 31.43	& 35.48	& 34.45	& 31.09	& \textbf{35.99}
                         \\
\hline
\end{tabular}
\end{center}
\end{table}


Unlike the D, H, A maps which represent quite different types of properties, the R, G and B channels all encode color information and are always treated as a whole to represent the color property. The DHA Separation has brought a significant improvement for the object detection. What if we separate the RGB channels (denoted as RGB Separation) like we did on the DHA?
The features from R, G, B networks are joint with that from DHA network to feed into the SVM.
And the result is shown in Column IV. We can see that the RGB Separation cannot boost the performance and the result mean $AP^b$ is even a little lower (31.43\% to 31.09\%). It implies that each property should be represented completely and then learned in an independent way.

Our final system lies in Colomn V. The UCM is added as another property map for CNN, besides the RGB, D, H, and A, to provide extra information. The performance is further boosted than that in Column II, and the UCM proves its effectivity to help object detection. And we get the best mean $AP^b$ 35.99\% on the \emph{val} set, exceeding the strongest baseline in Column I by 4.56\% (14.51\% relative).

\begin{table*}[!htbp]
\scriptsize
\caption{\textbf{Test set performance on NYUD2 for RGB-D object detection.} We compare the results of our final system with the three existing remarkable methods: RGBD DPM, RGB+D CNN, RGB+DHA CNN. The finetuning set for CNN is shown in the Column 2 and the SVM is trained on the \emph{trainval} set. All the results are $AP^b$ in percentage. See Section.~\ref{sec_result_nyud2_test}}
\label{tab_nyud2_test}
\begin{center}
\begin{tabular}{|c|c|p{0.4cm}|p{0.27cm}p{0.27cm}p{0.27cm}p{0.27cm}p{0.27cm}p{0.27cm}p{0.27cm}p{0.27cm}p{0.27cm}p{0.27cm}p{0.27cm}p{0.27cm}p{0.27cm}p{0.27cm}p{0.27cm}p{0.27cm}p{0.27cm}p{0.27cm}p{0.4cm}|}
\hline

\rotatebox{90}{method}
& \rotatebox{90}{\tabincell{c}{finutune\\set}}
& \rotatebox{90}{mean}
& \rotatebox{90}{bathtub} & \rotatebox{90}{bed}     & \rotatebox{90}{\tabincell{c}{book-\\shelf}}        & \rotatebox{90}{box}     & \rotatebox{90}{chair}
& \rotatebox{90}{counter} & \rotatebox{90}{desk}    & \rotatebox{90}{door}        & \rotatebox{90}{dresser} & \rotatebox{90}{\tabincell{c}{garbage-\\bin}}
& \rotatebox{90}{lamp}    & \rotatebox{90}{monitor} & \rotatebox{90}{\tabincell{c}{night-\\stand}} & \rotatebox{90}{pillow}  & \rotatebox{90}{sink}
& \rotatebox{90}{sofa}    & \rotatebox{90}{table}   & \rotatebox{90}{\tabincell{c}{televi-\\sion}}  & \rotatebox{90}{toilet}      \\
\hline

RGBD DPM
& -
& 23.9
& 19.3  & 56.0  & 17.5  & 0.6   & 23.5
& 24.0  & 6.2   & 9.5   & 16.4  & 26.7
& 26.7  & 34.9  & 32.6  & 20.7  & 22.8
& 34.2  & 17.2  & 19.5  & 45.1      \\ 
\hline

RGB+D CNN
& train
& 33.00	
& 39.97	& 64.57	& 36.38	& 1.35	& 42.04	
& 41.37	& 8.23	& 18.99	& 20.82	& 29.90	
& 34.71	& 42.69	& 27.45	& 32.68	& 37.26	
& 43.43	& 23.45	& 32.63	& 49.14		\\ 
\hline

RGB+DHA CNN
& train
& 36.14	
& 40.99	& 68.64	& 35.17	& 2.05	& 42.84	
& 44.46	& 13.28	& 21.91	& 29.43	& 30.28	
& 36.39	& 45.54	& 31.95	& 39.82	& 35.47	
& 49.44	& 24.64	& \textbf{40.25} & \textbf{54.17}
\\
\hline

Ours
& train
& 40.18	
& 47.68	& 73.53	& 38.77	& 2.49	& 49.30	
& 47.54	& 11.76	& 25.90	& 28.74	& 41.56	
& 36.12	& 56.75	& \textbf{47.27} & 42.04	& 45.61	
& 53.40	& \textbf{28.83} & 35.58	& 50.47
\\
\hline

RGB+D CNN
& trainval
& 36.11
& 28.67 & 70.07 & 37.90 & 4.18  & 46.47
& 41.52 & 13.03 & 21.20 & 26.40 & 35.36
& 37.74 & 53.06 & 36.87 & 36.59 & 43.30
& 46.84 & 23.38 & 31.12 & 52.37		\\
\hline

RGB+DHA CNN
& trainval
& 38.94
& \textbf{49.10} & 72.80 & 37.81 & 4.23  & 47.85
& \textbf{49.17} & \textbf{18.41} & 23.54 & \textbf{33.15} & 42.27
& 38.49 & 54.80 & 36.16 & 40.58 & 41.76
& 51.85 & 22.47 & 27.55 & 47.84
\\
\hline

Ours
& trainval
& \textbf{41.85}
& 45.50 & \textbf{74.97} & \textbf{42.86} & \textbf{5.23}  & \textbf{52.49}
& 47.49 & 15.94 & \textbf{27.25} & 29.98 & \textbf{45.53}
& \textbf{40.53} & \textbf{57.81} & 45.39 & \textbf{47.97} & \textbf{46.10}
& \textbf{54.63} & 27.08 & 37.30 & 51.16
\\
\hline

\end{tabular}
\end{center}
\end{table*}

\subsection{Performance comparison}
\label{sec_result_nyud2_test}
When it comes to the performance on \emph{test} set, we compare our final system with the existing remarkable methods: RGBD DPM, RGB+D CNN and RGB+DHA CNN.  The result is shown in Table~\ref{tab_nyud2_test}.

The Row 1 shows the result of RGBD DPM from~\cite{kim2013accurate, gupta2014learning}, which is the state-of-art method before the revival of CNN.
And the Row 2 and Row 3 are the results of RGB+D CNN and RGB+DHA CNN, which agree with~\cite{gupta2014learning} (with no data augmentation). Then the Row 4 gives the mean $AP^b$ of 40.18\% achieved by our system when finetuning the CNNs on the $train$ set. We improve the performance by 4.04\% over the best baseline (36.14\% to 40.18\%, 11.18\% relative).

Then we finetune the network again on the \emph{trainval} set. Thanks to more training data, the trained CNN gains higher generalization power. We get the performance shown in Row 5 to 7. Our system's result (Row 7) is still much better than the baselines (Row 5 and Row 6). Our final system improves the best mean $AP^b$ on the \emph{test} set of \textit{NYUD2} from 38.94\% to 41.85\%.

We have noticed that~\cite{gupta2015aligning} reported competitive performance with us by adding region features besides the bounding boxes. They expanded the system in~\cite{gupta2014learning} in a different way from us. Certainly we can also add regions features in our system. However, that's not the point of this paper. We have already gotten state-of-the-art performance even without region features.

\section{Conclusion}
\label{sec_conclusion}
In this paper, we addressed the problem of deeply exploiting the deep information for RGB-D object detection, and proposed a novel framework.
Specifically, we first derived more properties by mining the provided RGB and depth data.
Particularly, several properties that could be directly derived from the color/depth or pairs were adopted here, which included the geometry contour, horizontal disparity, height above ground, and angle with gravity.
Then we systematically investigated the fusion schemes of different properties under the CNN model. By the means of analysis and evaluation, it was recommended that the features encoding the different object properties should be learned independently and fused at the highest level, \ie not joint until passing into the classifier. Finally, we experimentally verified the effectiveness of the proposed method, which indeed achieved state-of-the-art performance on \textit{NYUD2}. And it was gained with no data augmentation or region features. Besides, we only considered the properties that could be computed in relatively straightforward methods. Exploring more useful properties is one of our future works, \eg, equipping more powerful sensors or developing more advanced algorithms for property derivation.

\section*{Acknowledgment}
This work is supported partially by the National High Technology Research and Development Program of China (863 program, 2014AA06A503), National Natural Science Foundation of China under Grant 61233003 and 61203256,  Natural Science Foundation of Anhui Province (1408085MF112), and the
Fundamental Research Funds for the Central Universities (WK3500000002 and WK3490000001). And we express heartfelt thanks for the generous donation of Tesla GPU K40m from the NVIDIA corporation.


\small
\begin{thebibliography}{10}\itemsep=-1pt

\bibitem{amfm_pami2011}
P.~Arbelaez, M.~Maire, C.~Fowlkes, and J.~Malik.
\newblock Contour detection and hierarchical image segmentation.
\newblock {\em IEEE Trans. Pattern Anal. Mach. Intell.}, 33(5):898--916, May
  2011.

\bibitem{arbelaez2014multiscale}
P.~Arbelaez, J.~Pont-Tuset, J.~Barron, F.~Marques, and J.~Malik.
\newblock Multiscale combinatorial grouping.
\newblock In {\em CVPR}, 2014.

\bibitem{couprie2013indoor}
C.~Couprie, C.~Farabet, L.~Najman, and Y.~LeCun.
\newblock Indoor semantic segmentation using depth information.
\newblock {\em CoRR}, abs/1301.3572, 2013.

\bibitem{dalal2005histograms}
N.~Dalal and B.~Triggs.
\newblock Histograms of oriented gradients for human detection.
\newblock In {\em CVPR}, 2005.

\bibitem{dollar2013structured}
P.~Doll{\'a}r and C.~L. Zitnick.
\newblock Structured forests for fast edge detection.
\newblock In {\em ICCV}, 2013.

\bibitem{einhauser2003does}
W.~Einhauser and P.~Konig.
\newblock Does luminance-contrast contribute to a saliency map for overt visual
  attention?
\newblock {\em Eur. J. Neurosci.}, 17:1089--1097, 2003.

\bibitem{erhan2014scalable}
D.~Erhan, C.~Szegedy, A.~Toshev, and D.~Anguelov.
\newblock Scalable object detection using deep neural networks.
\newblock In {\em CVPR}, 2014.

\bibitem{everingham2010pascal}
M.~Everingham, L.~Van~Gool, C.~K. Williams, J.~Winn, and A.~Zisserman.
\newblock The pascal visual object classes (voc) challenge.
\newblock {\em International journal of computer vision}, 88(2):303--338, 2010.

\bibitem{fan2008liblinear}
R.-E. Fan, K.-W. Chang, C.-J. Hsieh, X.-R. Wang, and C.-J. Lin.
\newblock Liblinear: A library for large linear classification.
\newblock {\em The Journal of Machine Learning Research}, 9:1871--1874, 2008.

\bibitem{felzenszwalb2010object}
P.~F. Felzenszwalb, R.~B. Girshick, D.~McAllester, and D.~Ramanan.
\newblock Object detection with discriminatively trained part-based models.
\newblock {\em Pattern Analysis and Machine Intelligence, IEEE Transactions
  on}, 32(9):1627--1645, 2010.

\bibitem{fu2015object}
H.~Fu, D.~Xu, S.~Lin, and J.~Liu.
\newblock Object-based rgbd image co-segmentation with mutex constraint.
\newblock In {\em CVPR}, 2015.

\bibitem{gazzaniga2013cognitive}
M.~Gazzaniga, R.~Ivry, and G.~Mangun.
\newblock {\em Cognitive Neuroscience: The Biology of the Mind (Fourth
  Edition)}.
\newblock W. W. Norton, 2013.

\bibitem{girshick2014rich}
R.~Girshick, J.~Donahue, T.~Darrell, and J.~Malik.
\newblock Rich feature hierarchies for accurate object detection and semantic
  segmentation.
\newblock In {\em CVPR}, 2014.

\bibitem{gupta2015aligning}
S.~Gupta, P.~Arbel{\'a}ez, R.~Girshick, and J.~Malik.
\newblock Aligning 3d models to rgb-d images of cluttered scenes.
\newblock In {\em CVPR}, 2015.

\bibitem{gupta2013perceptual}
S.~Gupta, P.~Arbelaez, and J.~Malik.
\newblock Perceptual organization and recognition of indoor scenes from rgb-d
  images.
\newblock In {\em CVPR}, 2013.

\bibitem{gupta2014learning}
S.~Gupta, R.~Girshick, P.~Arbel{\'a}ez, and J.~Malik.
\newblock Learning rich features from rgb-d images for object detection and
  segmentation.
\newblock In {\em ECCV}, 2014.

\bibitem{hariharan2014simultaneous}
B.~Hariharan, P.~Arbel{\'a}ez, R.~Girshick, and J.~Malik.
\newblock Simultaneous detection and segmentation.
\newblock In {\em ECCV}, 2014.

\bibitem{jia2014caffe}
Y.~Jia, E.~Shelhamer, J.~Donahue, S.~Karayev, J.~Long, R.~Girshick,
  S.~Guadarrama, and T.~Darrell.
\newblock Caffe: Convolutional architecture for fast feature embedding.
\newblock {\em arXiv preprint arXiv:1408.5093}, 2014.

\bibitem{kim2013accurate}
B.-s. Kim, S.~Xu, and S.~Savarese.
\newblock Accurate localization of 3d objects from rgb-d data using
  segmentation hypotheses.
\newblock In {\em CVPR}, 2013.

\bibitem{kong2015bilinear}
Y.~Kong and Y.~Fu.
\newblock Bilinear heterogeneous information machine for rgb-d action
  recognition.
\newblock In {\em CVPR}, 2015.

\bibitem{krahenbuhl2012efficient}
P.~Krahenbuhl and V.~Koltun.
\newblock Efficient inference in fully connected crfs with gaussian edge
  potentials.
\newblock In {\em NIPS}, 2011.

\bibitem{krizhevsky2012imagenet}
A.~Krizhevsky, I.~Sutskever, and G.~E. Hinton.
\newblock Imagenet classification with deep convolutional neural networks.
\newblock In {\em NIPS}, 2012.

\bibitem{long_shelhamer_fcn}
J.~Long, E.~Shelhamer, and T.~Darrell.
\newblock Fully convolutional networks for semantic segmentation.
\newblock In {\em CVPR}, 2015.

\bibitem{lowe2004distinctive}
D.~G. Lowe.
\newblock Distinctive image features from scale-invariant keypoints.
\newblock {\em International journal of computer vision}, 60(2):91--110, 2004.

\bibitem{Silberman:ECCV12}
P.~K. Nathan~Silberman, Derek~Hoiem and R.~Fergus.
\newblock Indoor segmentation and support inference from rgbd images.
\newblock In {\em ECCV}, 2012.

\bibitem{rowley1998neural}
H.~Rowley, S.~Baluja, T.~Kanade, et~al.
\newblock Neural network-based face detection.
\newblock {\em Pattern Analysis and Machine Intelligence, IEEE Transactions
  on}, 20(1):23--38, 1998.

\bibitem{sermanet2013pedestrian}
P.~Sermanet, K.~Kavukcuoglu, S.~Chintala, and Y.~LeCun.
\newblock Pedestrian detection with unsupervised multi-stage feature learning.
\newblock In {\em CVPR}, 2013.

\bibitem{shotton2013real}
J.~Shotton, T.~Sharp, A.~Kipman, A.~Fitzgibbon, M.~Finocchio, A.~Blake,
  M.~Cook, and R.~Moore.
\newblock Real-time human pose recognition in parts from single depth images.
\newblock {\em Communications of the ACM}, 56(1):116--124, 2013.

\bibitem{socher2012convolutional}
R.~Socher, B.~Huval, B.~Bath, C.~D. Manning, and A.~Y. Ng.
\newblock Convolutional-recursive deep learning for 3d object classification.
\newblock In {\em NIPS}, 2012.

\bibitem{szegedy2014going}
C.~Szegedy, W.~Liu, Y.~Jia, P.~Sermanet, S.~Reed, D.~Anguelov, D.~Erhan,
  V.~Vanhoucke, and A.~Rabinovich.
\newblock Going deeper with convolutions.
\newblock In {\em CVPR}, 2015.

\bibitem{szegedy2013deep}
C.~Szegedy, A.~Toshev, and D.~Erhan.
\newblock Deep neural networks for object detection.
\newblock In {\em NIPS}, 2013.

\bibitem{zeiler2014visualizing}
M.~D. Zeiler and R.~Fergus.
\newblock Visualizing and understanding convolutional networks.
\newblock In {\em ECCV}, 2014.

\end{thebibliography}

\end{document}